\documentclass[a4paper,11pt]{extarticle}

\usepackage[utf8]{inputenc}
\usepackage{amsmath,amssymb,mathrsfs,amsthm}
\usepackage[margin=1in]{geometry}

\usepackage{graphicx}
\usepackage{multirow}
\usepackage[export]{adjustbox}
\usepackage{bm}
\usepackage{float}
\usepackage{subcaption}
\usepackage[english]{babel}
\newtheorem*{prop}{Proposition}
\usepackage{algorithm}
\usepackage[noend]{algpseudocode}
\usepackage{comment}
\usepackage{xcolor}
\usepackage{arydshln}
\usepackage{tikz}
\usepackage{pgfplots}
\usetikzlibrary{spy, calc}
\usepgfplotslibrary{colorbrewer}
\usepgfplotslibrary{groupplots}

\usetikzlibrary{pgfplots.groupplots}
\usetikzlibrary{matrix,arrows,decorations.pathmorphing}
\usepgfplotslibrary{fillbetween}
\long\def\comment#1{}
\newcommand\vtheta{{\bm \theta}}

\usepackage{spy}
\usepackage{rays_defs_17}

\usepackage{hyperref}
\hypersetup{
    unicode=false,          % non-Latin characters in Acrobat‚Äö√Ñ√∂‚àö√ë‚àö¬•s bookmarks
    pdftoolbar=true,        % show Acrobat‚Äö√Ñ√∂‚àö√ë‚àö¬•s toolbar?
    pdfmenubar=true,        % show Acrobat‚Äö√Ñ√∂‚àö√ë‚àö¬•s menu?
    pdffitwindow=false,     % window fit to page when opened
    pdfstartview={FitH},    % fits the width of the page to the window
    pdfauthor={Youssef Mansour},     % author
    pdfsubject={Subject},   % subject \of the document
    pdfcreator={Youssef Mansour},   % creator of the document
    pdfproducer={Producer}, %!TEX encoding = UTF-8 Unicode: producer of the document
    pdfnewwindow=true,      % links in new window
    colorlinks=true,       % false: boxed links; true: colored links
    linkcolor=red,          % color of internal links
    citecolor=green,        % color of links to bibliography
    filecolor=magenta,      % color of file links
    urlcolor=cyan           % color of external links
}

\usepackage[
backend=bibtex,
url=false,isbn=false,doi=false,
hyperref=auto,
style=alphabetic,
natbib=true,
sorting=nty,
firstinits=true,
maxnames=2, maxbibnames=10,
]{biblatex}
\bibliography{./references.bib} 

\title{\textbf{Zero-Shot Noise2Noise: Efficient Image Denoising without any Data}}
\author{Youssef Mansour* and Reinhard Heckel*$^{,\dagger}$ \\
*School of Computation, Information and Technology, Technical University of Munich\\
*Munich Center for Machine Learning\\
$^\dagger$Dept. of Electrical and Computer Engineering, Rice University\\ }
\date{}
\begin{document}
\maketitle

\begin{abstract}
Recently, self-supervised neural networks have shown excellent image denoising performance. However, current dataset free methods are either computationally expensive, require a noise model, or have inadequate image quality. In this work we show that a simple 2-layer network, without any training data or knowledge of the noise distribution, can enable high-quality image denoising at low computational cost. Our approach is motivated by Noise2Noise and Neighbor2Neighbor and works well for denoising pixel-wise independent noise. Our experiments on artificial, real-world camera, and microscope noise show that our method termed ZS-N2N (Zero Shot Noise2Noise) often outperforms existing dataset-free methods at a reduced cost, making it suitable for use cases with scarce data availability and limited computational resources. A demo of our implementation including our code and hyperparameters can be found in the following \href{https://colab.research.google.com/drive/1i82nyizTdszyHkaHBuKPbWnTzao8HF9b?usp=sharing}{colab notebook}.
\end{abstract}

\section{Introduction}

Image denoising is the process of removing distortions from images, to enhance them visually and to reconstruct fine details. The latter is especially important for medical images, where fine details are necessary for an accurate diagnosis.  

Current state-of-the-art image denoising techniques rely on large data sets of clean-noisy image pairs and often consist of a neural network trained to map the noisy to the clean image. The drawbacks of dataset based methods are that data collection, even without ground truths, is expensive and time-consuming, and second, a network trained on dataset suffers from a performance drop if the test images come from a different distribution of images. These drawbacks motivate research in dataset-free methods.

All current zero-shot models are either suitable only for specific noise distributions and need previous knowledge of the noise level \cite{bm3d,anscombe}, require a lot of compute (time, memory, GPU) to denoise an image \cite{self2self}, have poor denoising quality \cite{dip}, or do not generalise to different noise distributions or levels \cite{noise2void, self2self}. 
We propose a method that builds on the recent Noise2Noise \cite{noise2noise} and Neighbour2Neighbour\cite{nb2nb} papers and aims to circumvent these issues to reach a good trade-off between denoising quality and computational resources. We make only minimal assumptions on the noise statistics (pixel-wise independence), and do not require training data. Our method does not require an explicit noise model, and is therefore suitable for various noise types and can be employed when the noise distribution or level are unknown. The only assumption we make about the noise is that it is unstructured and has zero mean.

In a nutshell, we convolve the noisy test image with two fixed filters, which yields two downsampled images. We next train a lightweight network with regularization to map one downsampled image to the other. Our strategy builds on the recent Noise2Noise \cite{noise2noise} and Neighbour2Neighbour\cite{nb2nb} papers, however we take those methods one step further by enabling denoising without any training data. Even with an extremely small network and without any training data, our method achieves good denoising quality and often even outperforms large networks trained on datasets.

The key attributes of our work are as follows: 

\begin{itemize}
\item \textbf{Compute.} Dataset free neural network based algorithms \cite{dip,self2self} require solving an optimization problem involving millions of parameters to denoise an image. The huge parameter count requires large memory storage, advanced GPUs, and long denoising times. In this work we show that our method, that utilizes a simple 2 layer network, with only 20k parameters, can often outperform networks with millions of parameters while reducing the computational cost significantly and being easily executable on a CPU.   
\item \textbf{Generalisation.} Existing zero-shot methods often to do not generalise well. For example, BM3D \cite{bm3d}, a classical denoising algorithm does not generalize well to non-Gaussian noise, and blind spot networks \cite{noise2void} \cite{self2self} (discussed later in detail) fail to denoise well in the regime of low noise level. Extensive experiments on different noise distributions and noise levels show that our proposed approach can generalise better to different conditions better than existing methods. 
\end{itemize}

\begin{figure}[t]
\begin{center}
\includegraphics[width=1\textwidth]{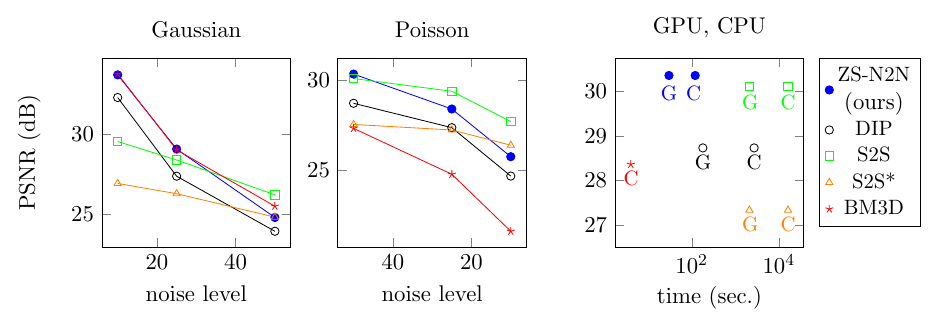}
\end{center}
\caption{
\label{fig:intro}
Left and middle plots: PSNR scores for Gaussian and Poission denoising for different noise levels. Note BM3D's poor performance on Poisson compared to Gaussian noise. Right plot: Time required in seconds to denoise one $256 \times 256$ colour image on CPU and GPU, tested on Poisson noise with $\lambda = 50$. Except for BM3D, all methods have shorter times on GPU. Only S2S in some cases outperforms our method, however it is about 100 times slower. S2S* denotes the ensemble free version of S2S.   
}
\end{figure}

In summary, our proposed method is dataset and noise model-free, and achieves a better trade-off between generalization, denoising quality, and computational resources compared to existing zero-shot methods, as displayed in Figure \ref{fig:intro}. We compare to the standard zero shot baselines, including BM3D, and the recent neural network-based algorithms DIP \cite{dip} and S2S \cite{self2self}. Only BM3D is faster than our method but achieves poor results on non-Gaussian noise. Only S2S sometimes outperforms our method, but is orders of magnitude slower, often fails on low noise levels \cite{kim2022zeroshot}, and requires ensembling to achieve acceptable performance. 

%%%%%%%%%%%%%%%%%%%%%%%%%%%%%%%%%%%%%%%%%%%%%%%%%%%%%%%%%%%%%%%%%%%%%%%%%%%%%%%%%%%%%%%%%%%%%

%%%%%%%%%%%%%%%%%%%%%%%%%%%%%%%%%%%%%%%%%%%%%%%%%%%%%%%%%%%%%%%%%%%%%%%%%%%%%%%%%%%%%%%%%%%%%

\section{Related Work}

\paragraph{Zero-Shot/ Dataset free Methods.} 
Our method is conceptually very similar to Noise2Fast \cite{noise2fast}, which also builds on Noise2Noise and Neighbour2Neighbour to achieve dataset-free denoising. 
However, Noise2Fast uses a relatively large network and requires an early stopping criterion. Our work improves on Noise2Fast by working with a consistency loss that alleviates the need to early stop, and using a much smaller network which saves compute. Specifically, our network is twelve times smaller and a forward pass through it is seven times faster. 
Our work utilizes a small 2-layer network and achieves competitive quality for image restoration. We show that on grayscale images, our method despite achieving similar scores to Noise2Fast \cite{noise2fast}, produces better-quality images. This is likely due to Noise2Fast dropping pixel values when downsampling, whereas our method keeps all information retained.  

Besides this work, classical non-learning-based methods, such as BM3D \cite{bm3d} and Anscombe \cite{anscombe}, work well for Gaussian and Poisson noise, respectively, and require the noise level as an input. %Therefore, developing new approaches that are less restrictive and make less assumptions on the noise distribution is advantageous.     

Another popular neural network-based technique is DIP (Deep Image Prior) \cite{dip} and its variants such as the Deep Decoder \cite{deepdecoder}. DIP builds on the fact that CNNs have an inductive bias towards natural images, in that they can fit natural images much faster than noise. Therefore, a network trained, with early stopping, to map a random input to the noisy image will denoise the image. The denoising performance of DIP is often poor, and is dependent on the number of training epochs, which is hard to determine in advance. 

Self2Self \cite{self2self} is another important method that achieves promising results. It utilizes the idea of the blind spot networks (reconstructing masked pixels) on a single image, but with dropout ensembling. However, this method is not computationally efficient, in that it requires long durations to denoise an image. According to the authors, it takes 1.2 hours to denoise one $256 \times 256$ image on a GPU. Compared to other blind spot networks, Self2Self achieves significantly better denoising scores, since it relies on ensembling, i.e., averaging the output of several networks. However, ensemble learning over smoothens the image, causing a loss of some details, despite the improvement in PSNR scores \cite{Ensemble}.

Similar to almost all supervised and self-supervised methods, both Self2Self and DIP use a UNet \cite{unet} or a variant of it as the backbone network in their architectures. A UNet typically has millions of parameters, making it unsuitable for compute limited applications. Our work departs from this scheme, by designing a shallow and simple network with few parameters.

\paragraph{Supervised methods} achieve state-of-the-art performance by training a network end-to-end to map a noisy image to a clean one. Networks that work well are CNNs~\cite{DnCNN, brooks_UnprocessingImagesLearned_2019}, vision  transformers~\cite{swintransformer}, or MLP based architectures \cite{Img2Img-Mixer, maxim}.

Noise2Noise \cite{noise2noise} yields excellent performance from training on two noisy images of the same static scene, without any ground truth images. Given that the noise has zero mean, training a network to map one noisy image to another noisy image of the same scene performs as well as mapping to the ground truth. While having access to a pair of noisy images of the same scene is in practice hard to achieve, the Noise2Noise method has motivated further research in self-supervised methods~\cite{nb2nb} that require only single noisy images. 

\paragraph{Self-supervised methods} are trained on datasets consisting of only noisy images. 

Noise2Void \cite{noise2void} and Noise2Self \cite{noise2self} are two blind spot prediction approaches for image denoising. Given a set of noisy images $\{\vy^i\}_1^n$, The idea is to minimize the loss $\frac{1}{n}\sum_{i=1}^{n} \\ \mathcal{L}(f_\vtheta(M^i(\vy^i)),\vy^i)$, where $\mathcal{L}$ is a loss function, $f_\vtheta$ is a network, and $M^i$ is an operator that masks some pixels, hence the name blind spot. Assuming that the neighbouring pixels of a clean image are highly correlated, and that the noise pixels are independent, a network trained to reconstruct a masked pixel, can only predict the signal value from the neighbouring visible pixels, but not the noise.

Blind spot networks require long training times and have low denoising quality. Probabilistic variations of such networks \cite{laine19, probab_noise2void} converge much faster, and use posterior mean estimation to achieve better quality. Those probabilistic variations of blind spot networks work  well for a given artificial noise model, but a significant performance drop was shown when using such methods to denoise real world camera noise, since the natural noise is  not well approximated by artificial noise \cite{nb2nb}. 

Recently, several works \cite{sure1, sure2, sure3} attempted to use Stein's unbiased risk estimator for Gaussian denoising. Such methods work well only for Gaussian noise and require the noise level to be known in advance. A more general framework is Noisier2Noise \cite{noisier2noise} which works for any known noise distribution. Noise is sampled and added to the noisy images to create noisier images. A network is then trained to map the noisier to the noisy images. However, working with double noisy images distorts the image even further, which degrades performance.   

The newly proposed Neighbour2Neighbour \cite{nb2nb} builds on the Noise2Noise \cite{noise2noise} method, where the assumptions are that the noise has zero mean and is pixel-wise independent. Neighbour2Neighbour extends Noise2Noise by enabling training without noisy image pairs. It does so by sub-sampling single noisy images to create pairs of noisy images, where Noise2Noise can be applied. Image sub-sampling is widely used in image processing tasks, such as compression \cite{compression} or as an augmentation technique to increase the training data.

%%%
\section{Method}

Our method builds on the Noise2Noise~\cite{noise2noise}, for training a network on pairs of noisy images, and the Neighbour2Neighbour (NB2NB) \cite{nb2nb}, which generates such pairs from a single noisy image. Our main idea is to generate a pair of noisy images from a single noisy image and train a small network only on this pair. We start with a brief summary of Noise2Noise and then introduce our method.

%%%

\subsection{Background: Noise2Noise and Neighbour2Neighbour}
\label{sec:noise2noise}

Supervised denoising methods are typically neural networks $f_\vtheta$ that map a noisy image $\vy$ to an estimate $f_\vtheta(\vy)$ of the clean image $\vx$. Supervised denoising methods are typically trained on pairs of clean images $\vx$ and noisy measurements $\vy = \vx + \ve$, where $\ve$ is noise. We refer to supervised denoising as Noise2Clean (N2C). 

Neural networks can also be trained on different noisy observations of the same clean image. Noise2Noise (N2N) \cite{noise2noise} assumes access to a set of pairs of noisy images $\vy_1 =\vx+\ve_1, \vy_2=\vx+\ve_2$, where $\ve_1,\ve_2$ are independent noise vectors. A network $f_\vtheta$ is then trained to minimize the empirical risk $\frac{1}{n} \sum_{i=1}^n \norm[2]{f_\vtheta(\vy_1^i) - \vy_2^i}^2$. 
This makes sense, since in expectation over such noisy instances, and assuming zero mean noise, training a network in a supervised manner to map a noisy image to another noisy image is equivalent to mapping it to a clean image i.e., 
\begin{equation}
\label{eqn:N2N=N2C}
    \argmin_{\vtheta} \EX{\norm[2]{f_{\vtheta}(\vy_1) - \vx}^2} = \argmin_{\vtheta} \EX{\norm[2]{f_{\vtheta}(\vy_1) - \vy_2}^2}.
\end{equation}

The proof is given in the supplementary material. 

In theory N2N training reaches the same performance as N2C training if the dataset is infinitely large. In practice, since the training set is limited in size, N2N falls slightly short of N2C. For example, N2N training with a UNet on 50k images gives a performance drop of only about 0.02 dB compared to N2C with a UNet.  

Despite the great performance of N2N, its usability is often limited, since it is difficult to obtain a pair of noisy images of the same static scene. For instance, the object being captured might be non-static, or the lighting conditions change rapidly. 

Neighbour2Neighbour (NB2NB) \cite{nb2nb} extends N2N and allows training only on a set of single noisy images, by sub-sampling a noisy image to create a pair of noisy images. Similar to N2N, NB2NB exhibits strong denoising performance when trained on many images.

\subsection{Zero-Shot Noise2Noise}

Our work extends Noise2Noise \cite{noise2noise} and Neighbour2Neighbour\cite{nb2nb} by enabling training on only one single noisy image. 
To avoid overfitting to the single image, we use a very 
%This is particularly difficult, since overfitting will most likely occur if a network is trained on only one image. We show that even without any data augmentation techniques, overfitting can be prevented by designing an extremely 
shallow network and an explicit regularization term.

Almost all self- or un-supervised denoising methods, including ours, rely on the premise that a clean natural image has different attributes than random noise. As shown in \cite{nb2nb}, a noisy image can be decomposed into a pair of downsampled images. %by averaging adjacent pixels patch-wise. 
Based on the premise that nearby pixels of a clean
image are highly correlated and often have similar values, while the noise pixels are unstructured and independent, the downsampled pair of noisy images has similar signal but independent noise. This pair can therefore serve as an approximation of two noisy observations of the same scene, where one observation is used as the input, and the other as the target, as in N2N.

Our approach is to first decompose the image into a pair of downsampled images and second train a lightweight network with regularization to map one downsampled image to the other. Applying the so-trained network to a noisy image yields the denoised image. We first explain how we generate the downsampled images, and then how we fit the network.

\paragraph{Image Pair Downsampler}

\begin{figure}[h]
\centering
{\includegraphics[width=0.5\textwidth]{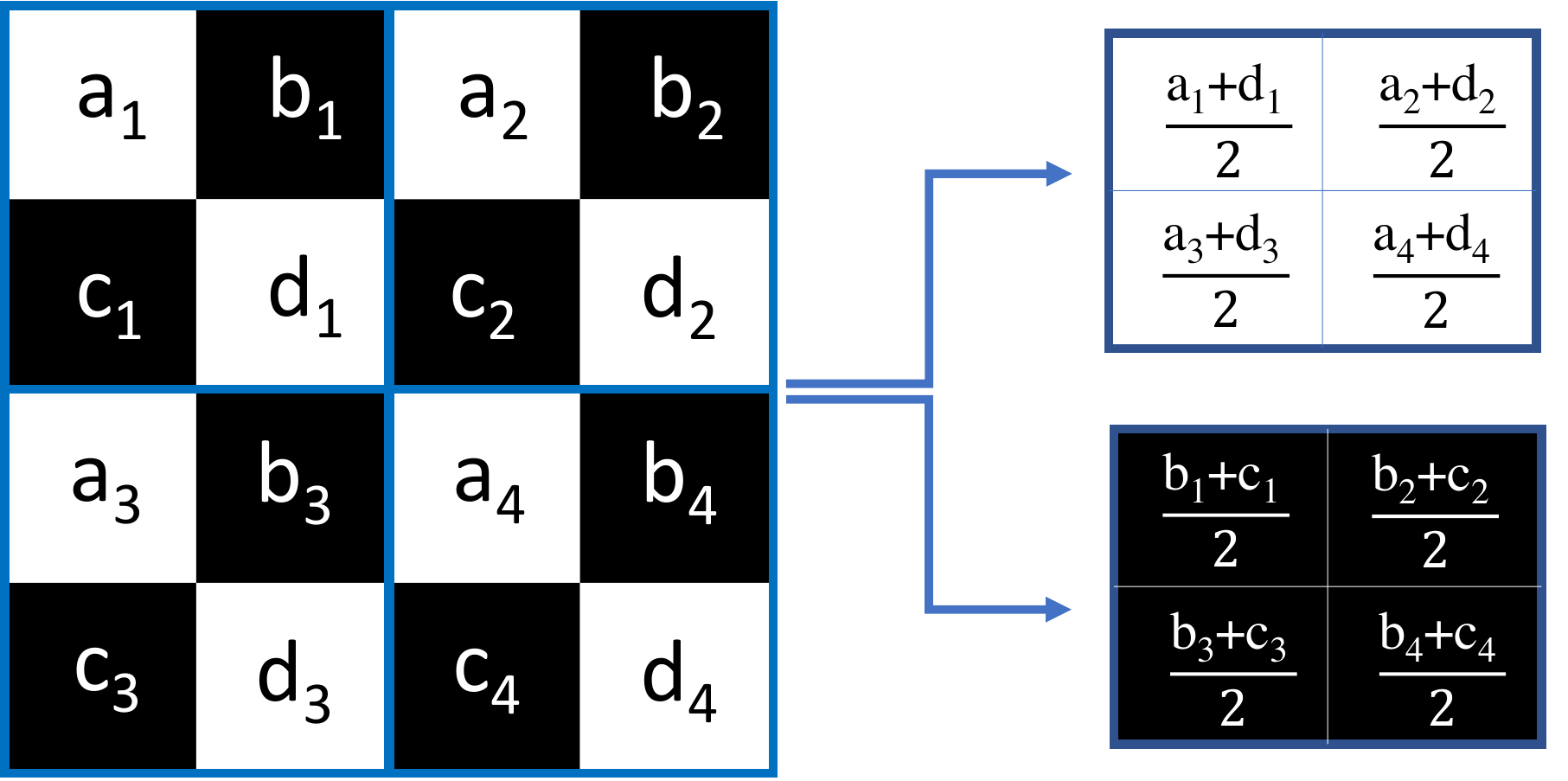}}
\caption{
\label{fig:img_pair_down}
The Image Pair Downsampler decomposes an image into two images of half the spatial resolution by averaging diagonal pixels of $2 \times 2$ non-overlapping patches. In the above example the input is a $4 \times 4$ image, and the output is two $2 \times 2$ images. 
}
\end{figure}

The pair downsampler takes as input an image 
 $\vy$ of size $H\times W\times C$ and generates two images $D_1(\vy)$ and $D_2(\vy)$, each of size $H/2\times W/2\times C$. 
The downsampler generates those images by dividing the image into non-overlapping patches of size $2\times2$, taking an average of the diagonal pixels of each patch and assigning it to the first low-resolution image, then the average of the anti-diagonal pixels and assigning it to the second low-resolution image. See Figure \ref{fig:img_pair_down} for an illustration of the pair downsampler. 

The downsampler is implemented with convolutions as follows. 
%The image pair downsampler generates two images of half the resolution by convolving the image with two fixed kernels of size $2\times2$ which downsample and average pixels. 
The first low-resolution image is obtained by applying a 2D convolution with stride two and fixed kernel $\bf{k}_1 = \begin{bmatrix}0 & 0.5 \\0.5 & 0\end{bmatrix}$ to the original image as 
$D_1(\vy) = \vy \circledast \bf{k}_1$, and the second image is obtained by applying a 2D convolution with stride two and fixed kernel  $\bf{k}_2 = \begin{bmatrix}0.5 & 0 \\0 & 0.5\end{bmatrix}$ to the original image as
$D_2(\vy) = \vy \circledast \bf{k}_2 $. The convolutions are implemented channel-wise and therefore the downsampling scheme is applicable to any arbitrary number of input channels. 
%Note that the kerne$\bf{k}_1$ and $\bf{k}_2$ are fixed, and do not change during training.

\paragraph{Zero-shot-image denoising method.}
Given a test image $\vy$ to denoise, our method is conceptually similar to first fitting a small image-to-image neural network $f_{\vtheta}$ to map the first downsampled image $D_1(\vy)$ to the second one, $D_2(\vy)$ by minimizing the loss
\begin{align}
\mc L(\vtheta) 
=
\norm[2]{f_{\vtheta}(D_1({\vy})) - D_2({\vy})}^2.
\end{align}
Once we fitted the network, we can apply it to the original noisy observation to estimate the denoised image as $\hat \vx = f_{\hat \vtheta}(\vy)$. 

However, our experiments showed that residual learning, a symmetric loss, and an additional consistency-enforcing term are critical for good performance. We next explain the elements of our loss function. 
In residual learning, the network is optimized to fit the noise instead of the image. The loss then becomes
\begin{equation}
    \mathcal{L}(\vtheta) =\norm[2]{D_1({\vy}) - f_{\vtheta}(D_1({\vy})) - D_2({\vy})}^2 . 
\end{equation}

Following \cite{simsiam}, where a symmetric loss was used in the context of self-supervised pretraining of a siamese network,  we additionally adopt a symmetric loss, which yields the residual loss: 
\begin{equation}
    \begin{aligned}
        \mathcal{L}_\mathrm{res.}(\vtheta) =\frac{1}{2} \Big(
         &\norm[2]{D_1({\vy}) - f_{\vtheta}(D_1({\vy})) - D_2({\vy})}^2 +\\
         &\norm[2]{D_2({\vy}) - f_{\vtheta}(D_2({\vy})) - D_1({\vy})}^2 \Big). 
    \end{aligned}
\end{equation}

In addition, we enforce consistency by ensuring that first denoising the image $\vy$ and then downsampling it, is similar to what we get when first downsampling $\vy$ and then denoising it, i.e., we consider a loss of the form:
\begin{equation}
 \mathcal{L}(\vtheta) = \norm[2]{D({\vy}) - f_{\vtheta}(D({\vy})) - D(\vy - f_{\vtheta}({\vy}))}^2.
\end{equation} 
 Again adopting a symmetric loss, the consistency loss becomes:
\begin{equation}
\begin{aligned}
    \mathcal{L}_\mathrm{cons.}(\vtheta) 
    = 
    \frac{1}{2} 
    \Big( &\norm[2]{D_1({\vy}) - f_{\vtheta}(D_1({\vy})) - D_1(\vy - f_{\vtheta}({\vy}))}^2 \\
    + &\norm[2]{D_2({\vy})-f_{\vtheta}(D_2({\vy})) - D_2(\vy-f_{\vtheta}({\vy}))}^2
    \Big).
\end{aligned}
\end{equation}

Note that for the residual loss, the network only has the downsampled images as input. Only in the consistency loss, the network gets to see the image in full spatial resolution. Including the consistency loss enables better denoising performance and helps to avoid overfitting. It can therefore be seen as a regularizing term.  

In summary, we minimize the loss $\mathcal{L}(\vtheta) = \mathcal{L}_\mathrm{res.}(\vtheta) + \mathcal{L}_\mathrm{cons.}(\vtheta)$ using gradient descent, which yields the network parameters $\hat \vtheta$. With those, we estimate the denoised image as $\hat \vx = \vy - f_{\hat \vtheta}(\vy)$. 
Note that only the network parameters $\vtheta$ are optimized during the gradient descent updates, since the downsampling operations $D_1$ and $D_2$ are fixed. Convergence typically requires 1k to 2k  iterations, which thanks to using a lightweight network takes less than half a minute on a GPU and around one minute on a CPU.

\paragraph{Network}
Many supervised and self-supervised methods use a relatively large network, often a UNet \cite{unet}. Instead, we use a very simple  two-layer image-to-image network. It consists of only two convolutional operators with kernel size 3 $\times$ 3 followed by one operator of 1$\times$1 convolutions. This network has about 20k parameters, which is small compared to typical denoising networks. An exact comparison of the network sizes can be found in section \ref{sec:comp_eff}. There are no normalization or pooling layers. The low parameter count and simple structure enables fast denoising even when deployed on a CPU. In the ablation studies we show that using a UNet instead of a lightweight network leads to overfitting and much worse denoising performance. 
%We denote our network by $f$ and its trainable parameters by $\vtheta$. 

\section{Experiments}

We compare our denoising algorithm (ZS-N2N) to several baselines. The baselines include dataset based methods, as well as other zero-shot methods. For the dataset based methods, we include both supervised (with clean images) and self-supervised (only noisy images) methods. We test all methods on artificial and real-world noise. We provide ablation studies in the supplementary material. 

The results highlight the dependency of dataset based methods on the dataset they are trained on and suggest that given a small training set, they are outperformed by dataset free ones. Furthermore, the experiments show that  methods based on noise models achieve good performance for the specific noise model, but do not generalise to other distributions.

Concerning the dataset and noise model free methods, our proposed method is either on par or better than other baselines on Gaussian, Poisson, and real world camera and microscope noise. Our method only falls short of Self2Self \cite{self2self} on high noise levels, however, it requires only $\frac{1}{200}$ of the denoising time of Self2Self and 2\% of it's memory. Moreover, Self2self's performance on low noise levels is insufficient. Therefore, considering denoising quality, generalistion, and computational resources, our method achieves a better trade-off compared to existing methods as shown in Figure \ref{fig:intro}.

\subsection{Baselines}
We compare to Noise2Clean (N2C) with a UNet, which is the current state-of-the-art denoising algorithm. There exits several other networks that perform on par with the UNet, such as DnCNN \cite{DnCNN} and RED30 \cite{red30}, but the UNet is orders of magnitude faster, since it is not very deep, and has a multi-resolution structure. The UNet is therefore the standard choice in all recent denoising papers \cite{noise2noise,noise2void,noisier2noise,nb2nb}.

For the self-supervised methods, we compare to Neighbour2Neighbour (NB2NB) \cite{nb2nb} and Noise2Void (N2V) \cite{noise2void}. We exclude the methods that require an explicit noise model, such as \cite{laine19,noisier2noise,sure1,sure3}, since these methods work well on synthetic denoising tasks for the given noise distribution, but fail to generalize to unknown noise distributions or real-world noise \cite{nb2nb, rrs}. This is due to the fact that the synthetic noise is insufficient for simulating real camera noise, which is signal-dependent and substantially altered by the camera's imaging system. 

Regarding the zero-shot methods, which are most similar to ours, we compare to the deep learning based algorithms: DIP \cite{dip} and Self2Self (S2S) \cite{self2self}, and also to the classical algorithm: BM3D \cite{bm3d}. % and Anscombe \cite{anscombe}. 
Note that apart of our method (and BM3D), all baselines use a U-Net or a variation of it as a denoising backbone. 

The performance of DIP is very sensitive to the number of gradient descent steps. We used the ground truth images to determine the best early stopping iteration. The DIP results can therefore be seen as an over optimistic performance of the method. For a fair comparison, we report the results of the best performing model for the other baselines. A comparison of the sensitivity of the methods to the number of optimization steps can be found in the supplementary material.    

The original implementation of S2S uses an ensemble of multiple networks, i.e, averaging the outputs of several networks. All other baselines do not utilize ensembling or averaging. For a fair comparison, we additionally report the results of S2S without any ensembling, which we denote by S2S*. S2S denotes the original implementation with an ensemble of 50 networks. 

\subsection{Synthetic Noise}
The dataset based methods (N2C, NB2NB, N2V) are trained on 500 colour images from ImageNet \cite{imagenet}. All methods are tested on the Kodak24 \footnote{\url{http://r0k.us/graphics/kodak/}} and McMaster18 \cite{McMaster} datasets. All training and test images are center-cropped to patches of size 256 $\times$ 256.

We examine Gaussian and Poisson noise with noise levels $\sigma$ and $\lambda$ respectively. We consider the fixed noise levels $\sigma,\lambda$= 10, 25, 50. The $\sigma$ values for Gaussian noise correspond to pixel values in the interval [0,255], while the $\lambda$ values for Poisson noise correspond to values in the interval [0,1].

For the dataset based methods, we also consider blind denoising during training with the range of noise levels $\sigma,\lambda \in [10,50]$. During training, a $\sigma,\lambda$ value is sampled uniformly from the given range for each image in each training epoch, unlike the fixed noise levels, where all training images are contaminated with the same noise level. Blind denoising is what is used in practice, since an exact noise level is typically not given, but rather a range of noise levels. 

In table \ref{tab:artificial_results}, we present the denoising performance of the different methods. For the dataset based methods, $\sigma,\lambda$ is known, denotes that the network trained on that exact noise level is used for testing, while unknown denotes the blind denoising, where the network trained on the range of noise levels [10,50] is used for testing. BM3D requires as input the value of the noise level. For Gaussian denoising the known $\sigma$ value was used, while for Possion denoising the noise level was estimated using the method in \cite{sigma_est}.  Note that ZS-N2N, DIP, and S2S do not utilize any prior information on the noise distribution or level. 

\begin{table*}[t]
\centering
\begin{tabular}{ |c|c c c|c c c|c c c| } 
\hline
Noise&\multicolumn{3}{c|}{Method} & \multicolumn{3}{c|}{Kodak24}& \multicolumn{3}{c|}{McMaster18}\\
\hline
\hline
\multirow{12}{*}{Gaussian} &&
 & $\sigma$ known? &$\sigma=10$ &$\sigma=25$&$\sigma=50$&$\sigma=10$ &$\sigma=25$&$\sigma=50$\\
\cline{4-10}
&\multirow{6}{*}{\rotatebox{90}{dataset-based}}&\multirow{2}{*}{\emph{N2C}} &yes&33.45&28.27&25.47&33.03&28.46&\bf{25.86}\\
&&&no&32.16&28.18&24.45&31.97&28.26&24.78\\ 

&&\multirow{2}{*}{\emph{NB2NB}}&yes&33.01&27.90&25.02&32.63&28.01&\underline{25.25}\\
&&&no&31.79&27.80&24.15&31.19&27.85&23.95\\ 

&&\multirow{2}{*}{\emph{N2V}} 
&yes&30.19&26.21&24.07&   30.95&26.50&23.94\\
&&&no&28.95&26.03&23.19&   29.64&26.31&22.67\\ 

\cline{2-10}
&\multirow{5}{*}{\rotatebox{90}{dataset-free}}&ZS-N2N (ours) & - &\underline{33.69} & \bf{29.07} & 24.81&  \underline{34.21}&\underline{28.80}&24.02\\

&&DIP & - & 32.28 & 27.38 & 23.95&33.07&27.61&23.03\\

&&S2S & - &29.54 & 28.39 &\bf{26.22}&30.78& 28.71 &25.03 \\

&&S2S* & - &26.93 & 26.29 &24.83 &27.64&26.48&23.79\\

&&BM3D &yes&\bf{33.74}&\underline{29.02}&\underline{25.51}&\bf{34.51}&\bf{29.21}&24.51\\
\hline
\hline
\multirow{12}{*}{Poisson} &&
  & $\lambda$ known? &$\lambda=50$ &$\lambda=25$&$\lambda=10$&$\lambda=50$ &$\lambda=25$&$\lambda=10$\\
\cline{4-10}
&\multirow{6}{*}{\rotatebox{90}{dataset-based}}&\multirow{2}{*}{\emph{N2C}} &yes&\underline{29.42}&27.49&\underline{26.25}&\underline{29.89}&28.20&26.42\\
&&&no&28.92&27.14&23.13 & 28.62&27.51& 24.32\\ 

&&\multirow{2}{*}{\emph{NB2NB}}
&yes&29.19&27.01&25.71    &29.41&27.79&25.95\\
&&&no&  28.53&26.88&23.60    &28.03&27.66&24.58\\ 

&&\multirow{2}{*}{\emph{N2V}} 
&yes  &27.73&25.55&23.77   &27.86&25.65&23.47\\
&&&no   &27.04&25.28&21.93   & 26.34&25.52&22.07\\ 
\cline{2-10}
&\multirow{5}{*}{\rotatebox{90}{dataset-free}}&ZS-N2N (ours) & - & \bf{29.45}& \underline{27.52}&24.92 & \bf{30.36}&\underline{28.41}&25.75\\

&&DIP & - & 27.51 & 25.84 &23.81 & 28.73&27.37&24.67\\

&&S2S & - & 28.89 & \bf{28.31} &\bf{27.29}&  \underline{30.11} & \bf{29.40} & \bf{27.71} \\

&&S2S* & -&26.75&26.40&25.63& 27.55& 27.24 & 26.39  \\

&&BM3D &no&28.36&26.58&24.20&27.33&24.77&21.59\\
\hline
\end{tabular}
\caption{\label{tab:artificial_results} 
PSNR scores in dB for Gaussian and Poisson denoising. Best result is in \textbf{bold}, second best result is \underline{underlined}. The dataset based methods are \emph{italicized}. Note DIP's mediocre scores and BM3D's performance drop between Gaussian and Poission noise. S2S has significantly lower scores in low noise as seen with $\sigma = 10$ and its ensemble free version S2S* has inadequate performance. Denoised samples can be found in the supplementary material. }
\end{table*}

As seen from the results, the dataset based methods often fall slightly short of the dataset free methods. This is due to the fact that they were only trained on 500 images, whereas they reach good performance when trained on larger datasets. In the supplementary material, we show that when N2C is trained on 4000 images, it outperforms all other baselines and its performance can keep improving with more training data. Another drawback of dataset based methods is that they are sensitive to the data they are trained on. They experience a performance drop when trained on a range of noise levels as opposed to a specific noise level as the test set. 

Regarding the zero-shot methods, DIP exhibited worse scores in all simulations. BM3D is tailored to work well for Gaussian denoising, where the exact noise variance is known and required as input. However, its performance dropped for Poisson noise, where the noise level was estimated. 

ZS-N2N and S2S do not rely on a specific noise model and therefore work consistently well for both Gaussian and Poisson noise. However, S2S suffers from at least two drawbacks. The first is it heavily relies on ensembling to achieve good scores as seen by comparing the results of S2S with S2S*. Despite improving the scores, ensembling oversmoothens the image causing a loss in some visual features \cite{Ensemble}. Note that all other baselines are ensemble free. The second drawback is that it performs worse than all other baselines on low noise levels, as seen in the Gaussian noise with $\sigma=10$.

Considering that DIP performs poorly, that BM3D only works well for Gaussian noise, and that S2S's performance without ensembling and on low noise levels is unsatisfactory, our method, ZS-N2N is the only dataset free denoising algorithm that performs well on different noise distributions and levels.

\subsection{Real-World Noise}
\paragraph{Camera noise:} %We next assess the baselines on real-world camera noise. 
Following \cite{self2self}, we evaluate on the PolyU dataset \cite{polyu} which consists of high-resolution images from various scenes captured by 5 cameras from the 3 leading brands of cameras: Canon, Nikon, and Sony. We also consider the SIDD \cite{sidd}, which consists of images captured by several smartphone cameras under different lighting conditions and noise patterns. 

Since the computational cost for running S2S is high, we randomly choose 20 images from both datasets to test on. The SIDD validation set has images of size $256 \times 256$. For consistency, we center-crop the PolyU images to patches of size $256 \times 256$. The results are shown in table \ref{tab:natural_results}. All methods perform similarly except for BM3D and the ensemble free version of S2S, which exhibit a notable performance drop. 

\begin{table}[H]
\centering
\begin{tabular}{ |c|c|c|c|c|c| } 
\hline
Dataset & ZS-N2N & DIP & S2S & S2S* & BM3D\\
\hline
PolyU & 36.92&\bf{37.07} &\underline{37.01}& 33.12&36.11 \\
\hline
SIDD &\underline{34.07}&\bf{34.31} &  33.98& 30.77 &28.19 \\
\hline
\end{tabular}
\caption{\label{tab:natural_results} Denoising PSNR in dB on real world camera noise.  
}
\end{table}

\paragraph{Microscope noise:} We additionally evaluate on the Fluorescence Microscopy dataset \cite{fmd}, which contains real grayscale fluorescence images obtained with commercial confocal, two-photon, and wide-field microscopes and representative biological samples such as cells, zebrafish, and mouse brain tissues. We pick random images from the test set to test on. We also compare to Noise2Fast (N2F) \cite{noise2fast}, for which code for denoising grayscale is available. The results are depicted in table \ref{tab:fmd_results}.

\begin{table}[H]
\centering
\begin{tabular}{|c|c|c|c|c|c|c|}
\hline
Image         & ZS-N2N & DIP & S2S & S2S* & BM3D & N2F  \\
\hline
Photon BPAE   & 30.73 & 29.22& \underline{30.90} & 29.49& 27.19 & \bf{30.93}    \\
Photon Mice   & \underline{31.42}& 30.01& \bf{31.51}&29.99 &29.48& 31.07    \\
Confocal BPAE &\underline{35.85}& 35.51& 31.01 & 29.54&33.23 & \bf{36.01}    \\
\hline
Average       &\bf{32.67} &\underline{31.58}& 31.14 &29.67& 29.97&\bf{32.67}     \\
\hline
\end{tabular}
\caption{\label{tab:fmd_results} PSNR in dB on real world microscope noise.}
\end{table}

\begin{figure*}[h]\centering
\captionsetup[subfigure]{labelformat=empty}
\begin{tikzpicture}[zoomboxarray, zoomboxes below, zoomboxarray inner gap=0.1cm, zoomboxarray columns=2, zoomboxarray rows=1]
    \node [image node]{ \includegraphics[width=0.21\textwidth]{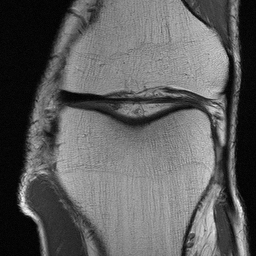}};
    \node at (1.7,-0.25) {Clean};
    \zoombox[magnification=3, color code = red]{0.55,0.6}
    \zoombox[magnification=2, color code = yellow]{0.2,0.5}
    %\zoombox[magnification=3, color code = blue]{0.75,0.18}
\end{tikzpicture}
\hfill
\begin{tikzpicture}[zoomboxarray, zoomboxes below, zoomboxarray inner gap=0.1cm, zoomboxarray columns=2, zoomboxarray rows=1]
    \node [image node] { \includegraphics[width=0.21\textwidth]{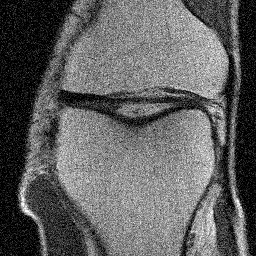}};
    \node at (1.7,-0.25) {Noisy 20.7 dB};
    \zoombox[magnification=3, color code = red]{0.55,0.6}
    \zoombox[magnification=2, color code = yellow]{0.2,0.5}
    %\zoombox[magnification=3, color code = blue]{0.75,0.18}
\end{tikzpicture}
\hfill
\begin{tikzpicture}[zoomboxarray, zoomboxes below, zoomboxarray inner gap=0.1cm, zoomboxarray columns=2, zoomboxarray rows=1]
    \node [image node] { \includegraphics[width=0.21\textwidth]{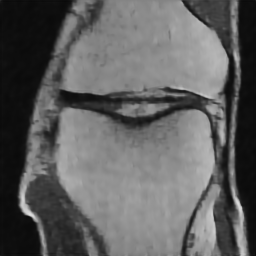}};
    \node at (1.7,-0.25) {Noise2Fast 28.0 dB};
    \zoombox[magnification=3, color code = red]{0.55,0.6}
    \zoombox[magnification=2, color code = yellow]{0.2,0.5}
    %\zoombox[magnification=3, color code = blue]{0.75,0.18}
\end{tikzpicture}
\hfill
\begin{tikzpicture}[zoomboxarray, zoomboxes below, zoomboxarray inner gap=0.1cm, zoomboxarray columns=2, zoomboxarray rows=1]
    \node [image node] { \includegraphics[width=0.21\textwidth]{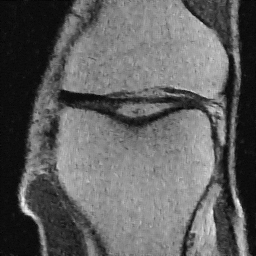}};
    \node at (1.7,-0.25) {Ours 27.8 dB};
    \zoombox[magnification=3, color code = red]{0.55,0.6}
    \zoombox[magnification=2, color code = yellow]{0.2,0.5}
    %\zoombox[magnification=3, color code = blue]{0.75,0.18}
\end{tikzpicture}
\caption{ \label{fig:mri} Visual comparison between our method and Noise2Fast for denoising Gaussian noise on a knee MRI. Both methods achieve similar PSNR, but notice how the center and left edge are blurry and oversmooth in Noise2Fast. Our method produces a sharper image with less loss of details.}
\end{figure*}
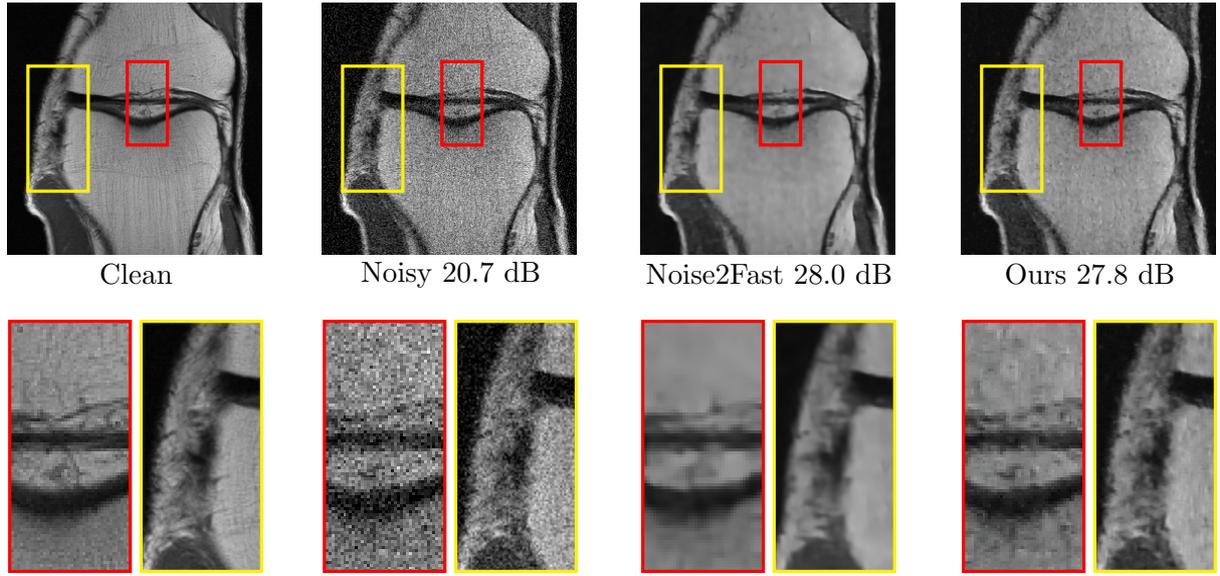

Our method and Noise2Fast achieve similar scores and slightly outperform the other baselines. Despite the similarity in scores, when inspecting the denoised images visually, we see differences: Our method produces visually sharper images and preserves slightly more details, while the Noise2Fast images are relatively smooth. This is most noticeable on images with fine details, such as MRI images, see Figure~\ref{fig:mri} for a knee image from the fastMRI dataset \cite{fastmri}. The blurriness in the Noise2Fast images is likely due to the downsampling scheme used, which drops some pixel values, and the ensembling performed to obtain the final image estimate, which oversmoothens the image \cite{Ensemble}. Our method, on the other hand, preserves all pixel values during downsampling, and is ensemble free.

\subsection{Computational Efficiency}
\label{sec:comp_eff}
%In the previous section we compared the performance, i.e., denoising accuracy, of the different methods. 
In this section we focus on the computational efficiency. We consider the denoising time and the memory requirements represented by the number of network parameters. Since in some applications a GPU is not available \cite{delbracio_PolyblurRemovingMild_2021}, we additionally consider the denoising time on a CPU. The GPU tested is Quadro RTX 6000 and the CPU is Intel Core i9-9940X 3.30GHz.

In table \ref{tab:comp_eff} we display the time required to denoise one colour image of size $256 \times 256$ at inference, as well as the total number of trainable parameters of a model. The dataset based methods are trained for long durations, but after training, the network parameters are fixed, and inference is almost instantaneous, since it is just a forward pass through the model. The time taken for denoising is therefore negligible compared to the zero-shot methods, whose parameters are optimized for each test image separately.  

In the original implementation of S2S, the authors report a denoising time of 1.2 hours for a $256 \times 256 $ colour image on GPU. However, we noticed that only half of the gradient update iterations are needed for convergence. We therefore report only half of their GPU time.

Concerning the denoising time, dataset based methods are the fastest, since a forward pass through a fixed network requires only milli seconds. 
%However, they are memory inefficient, in that they require networks with millions of parameters. 
Regarding the deep learning based zero-shot methods, ZS-N2N is significantly more computationally efficient.  Specifically, on CPU it is 200 times and 35 times faster than S2S and DIP respectively and has only 2\% and 1\% of their memory requirements. Only the classical BM3D is computationally more efficient than ZS-N2N.

\begin{table*}
\centering
\begin{tabular}{ |c||c|c|c||c|c|c|c| } 
\hline
Method & N2C & NB2NB & N2V & ZS-N2N & DIP & S2S & BM3D\\
\hline
GPU time & - & - & - &20 sec. & 3 min. & 35 min. & 4 sec. \\
CPU time & - & - & - &80 sec. &  45 min. &  4.5 hr. & 4 sec. \\
Network size & 3.3M & 1.3M& 2.2M &22k &2.2M&1M &- \\
\hline
\end{tabular}
\caption{\label{tab:comp_eff}Computational Resources. \textbf{First and Second Rows:} Time taken to denoise one image on average on GPU and CPU. The time for the dataset based methods is discarded, since it is negligible. BM3D does not benefit from the GPU, as there is no optimization involved. \textbf{Bottom Row:} Number of parameters of a network.}
\end{table*}

\subsection{Discussion}

Dataset based methods typically achieve state-of-the-art results but our experiments manifested two of their shortcomings: They don't perform well when trained on small datasets, and the performance drops when the test data differs from the training data, as seen by varying the noise levels. 
This highlights the importance of dataset free denoising algorithms.

Methods that rely on an explicit model of the noise distribution such as Noisier2Noise \cite{noisier2noise} and Anscombe \cite{anscombe} or those tailored to work well for specific distributions such as BM3D, do not generalize well to other distributions. Their performance therefore degrades when the noise distribution is unknown, or the noise level must be estimated. This has been manifested by BM3D's competitive performance on Gaussian noise, but its failure to keep up with the other baselines on Poission and real world noise. These findings highlight the advantage of noise model free techniques. 

Regarding the three dataset free and noise model free methods considered, DIP was often lagging behind S2S and ZS-N2N, despite using the ground truths to find the best possible early stopping iteration. S2S's performance without ensembling is inadequate, and even with ensembling, it does not work well on low noise levels. Moreover, it requires more than 0.5 hours to denoise an image on a GPU and 4.5 hours on a CPU. 

Except for ZS-N2N, all deep learning based baselines have millions of parameters, making them computationally expensive. Considering ZS-N2N's ability to generalize to various denoising conditions with relatively fast denoising time, very few parameters, and CPU compatibility, we can conclude that it offers a good trade-off between denoising quality and computational resources.

\section{Conclusion}
We proposed a novel zero-shot image denoising algorithm that does not require any training examples or knowledge of the noise model or level. Our work uses a simple 2-layer network, and allows denoising in a relatively short period of time even when executed without a GPU. The method can perform well on simulated noise as well as real-world camera and microscope noise, and achieves a good trade-off between generalization, denoising quality and computational resources compared to existing dataset free methods.

\section*{Acknowledgements}
The authors are supported by the Institute of Advanced Studies at the Technical University of Munich, the Deutsche Forschungsgemeinschaft (DFG, German Research Foundation) - 456465471, 464123524, the German Federal Ministry of Education and Research, and the Bavarian State Ministry for Science and the Arts. The authors of this work take full responsibility for its content.

%%%%%%%%% REFERENCES
{ 
\AtNextBibliography{\small} 
\printbibliography
}

\newpage
\appendix

%\title{Supplementary Material}
%\maketitle

\section*{\centering{Supplementary Material}}

\section{Ablation Studies}
In this section we provide additional experiments and discuss a few variants of our proposed approach to show which elements are essential for good performance. Unless otherwise mentioned, the ablation studies are conducted on the Kodak24 dataset contaminated with Gaussian noise of $\sigma = 25$.

\paragraph{Loss function} We study 3 variations of our proposed loss function, namely without the symmetric loss, without the consistency loss, and without the residual loss. The results are displayed in table \ref{tab:ablate_loss_fnc}. The symmetric loss offers minor improvement to the method's performance, where as the consistency loss has a more significant impact. However the residual loss is necessary, since without it the network just learns the identity mapping.  

\begin{table}[H]
\centering
\begin{tabular}{ |c|c|c|c| } 
\hline
 Default & w/o symmetric & w/o consistency & w/o residual\\
\hline
29.07& 28.65 & 28.01 & 17.93  \\
\hline
\end{tabular}
\caption{\label{tab:ablate_loss_fnc} Denoising PSNR in dB of ablated versions of the loss function. }
\end{table}

\paragraph{Network size} We saw that compared to deep learning based algorithms, ZS-N2N has few network parameters. In this section, we show that, perhaps surprisingly, even with much fewer parameters, ZS-N2N can still perform well. Moreover, we show that denoising with a UNet fails. This is most likely due to overfitting, as only a single test image is used for training. The results are depicted in table \ref{tab:ablate_net_size}. Even with a network as small as 500 parameters, ZS-N2N outperforms DIP that has 2 million parameters. 

\begin{table}[H]
\centering
\begin{tabular}{ |c||c|c|c|c|c|c|c| } 
\hline
 Network size & UNet (3.3M) & Default (22k) &4k& 2k & 1k & 500 \\
\hline
PSNR & 21.01 & 29.07& 28.66& 28.28  & 28.07 & 27.71 \\
\hline
\end{tabular}
\caption{\label{tab:ablate_net_size} Denoising PSNR in dB for reducing the number of parameters of the ZS-N2N network. The network size was reduced by decreasing the number of channels in the hidden layers.}
\end{table}

\paragraph{Data scaling}  In the previous experiments, the dataset based methods were trained on only 500 images, and therefore exhibited slightly worse performance than dataset free ones. In this section, we unveil the potential of the supervised Noise2Clean by additionally training on 4000 and 10000 images. As before, a UNet with 3.3M parameters is trained on ImageNet images.

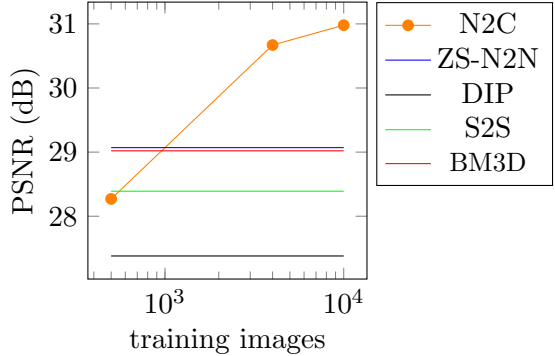
\begin{figure}[h]
\begin{center}
\begin{tikzpicture}
\begin{groupplot}[
xmode=log,
legend style={at={(1.35,1)},anchor=north}, 
legend style={cells={align=center}},
y label style={at={(axis description cs:0.20,0.5)},anchor=south},
group style={group size=1 by 1, horizontal sep=0.3cm, 
yticklabels at=edge left,
xticklabels at=edge bottom,
},
width=0.33\textwidth,height=0.33\textwidth, 
scaled x ticks=true,
every x tick label/.append style={alias=XTick,inner xsep=0pt},
every x tick scale label/.style={at=(XTick.base east),anchor=base west},
]
\nextgroupplot[xlabel = {training images},ylabel={PSNR (dB)}]
\addplot[mark=*,color=orange] coordinates { (500,28.27) (4000,30.67) (10000,30.98)};
\addlegendentry{{N2C}}
\addplot[mark=none,color=blue] coordinates {(500,29.07)  (10000,29.07) };
\addlegendentry{{ZS-N2N}}
\addplot[mark=none,color=black] coordinates {(500,27.38)  (10000,27.38) };
\addlegendentry{{DIP}}
\addplot[mark=none,color=green] coordinates {(500,28.39) (10000,28.39) };
\addlegendentry{{ S2S}}
\addplot[mark=none, color=red] coordinates {(500,29.02) (10000,29.02)};
\addlegendentry{{\small BM3D}}
\end{groupplot}
\end{tikzpicture}
\end{center}
\caption{
\label{fig:ablate_data_scale}
Denoising performance of Noise2Clean as the training set is scaled to larger sizes. Note that the performance of the dataset free methods is constant, since they do not make use of any training data.  
}
\end{figure}

The results are shown in figure \ref{fig:ablate_data_scale}. Already at 4000 training images, N2C significantly outperforms all other dataset free methods. These findings coincide with the results in the literature, that supervised dataset based methods achieve state-of-the-art results, given enough training data and similarity between the training and test sets.

\paragraph{Performance vs optimization iterations} 

\begin{figure}[ht]
\begin{center}
\includegraphics[width=1\textwidth]{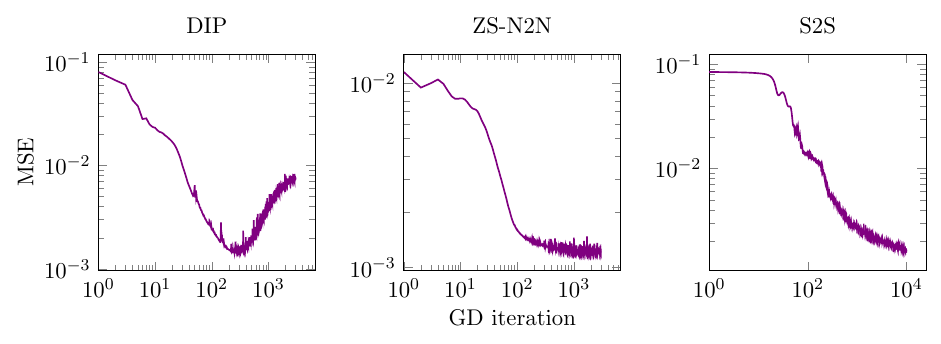}
\end{center}
\caption{
\label{fig:ablate_perf_iter}
Denoising one image from the Kodak24 dataset. The y-axis is the MSE between the clean image and the output of the network (denoised image). The x-axis is the gradient descent iterations.
}
\end{figure}

DIP's performance is sensitive to the number of gradient descent iterations. The optimal early stopping point for DIP varies according to noise type and level, which makes it hard to determine in advance. However, unlike DIP and similar to S2S, ZS-N2N's performance only improves with the optimization steps. An example is shown in figure \ref{fig:ablate_perf_iter}. This enables ZS-N2N to be deployed in various use cases with no manual interference or fine tuning.

\section{Appendix}

\paragraph{Weaknesses and Limitations}

ZS-N2N exhibits strong denoising performance and outperforms or is on par with other baselines in low and moderate noise levels. However, in high noise levels such as Gaussian noise with $\sigma = 50$, or Poission noise with $\lambda = 10$, a performance drop is noticed. ZS-N2N's performance in high noise levels is still better than DIP and BM3D, but worse than S2S. Nevertheless, ZS-N2N generalizes better than S2S, since it's performance in the high noise regime is acceptable and above other baselines, while S2S's performance in the low noise regime is poor. Moreover, even with high noise, one could still choose ZS-N2N over S2S if only limited compute is available or short denoising duration is required.  

Another weakness that ZS-N2N shares with all dataset free methods, is that they do not make use of training data. Therefore in use cases where abundant data is available that is similar to the test data, dataset based methods will significantly outperform zero-shot methods as seen in the ablation studies.

\paragraph{Proof of equation \ref{eqn:N2N=N2C}: }

\begin{prop}
Let $\norm[2]{\cdot}^2$ denote the squared $\mathcal{L}_2$ norm, and $\vtheta$ the trainable parameters of a network $f$. Let $\vy_1$ and $\vy_2$ be two noisy fixed observations of the same clean image $\vx$, i.e. $\vy_1 = \vx + \ve_1$ 
and $\vy_2 = \vx + \ve_2$, where $\ve_i$ is noise. Given that the $\ve_i$ are independent, and $\EX{\ve}=0$, the optimization problem w.r.t the MSE of Noise2Noise is the same as that of Noise2Clean. 
% Let $D$ denote a dataset of noisy image pairs $\{\vy_1^i, \vy_2^i \}_1^n$ with cardinality $\lvert D \rvert =n$. , and  $n\to\infty$
\end{prop}

Proof:

\begin{align*}
    \vtheta_{\mathrm{N2C}} &= \argmin_{\vtheta} \EX{\norm[2]{f_{\vtheta}(\vy_1) - \vx}^2} \\
    &= \argmin_{\vtheta}\EX{\norm[2]{f_{\vtheta}(\vy_1)}^2 - 2 \transp{\vx} f_{\vtheta}(\vy_1)}.\\
     \vtheta_{\mathrm{N2N}}&= \argmin_{\vtheta} \EX{\norm[2]{f_{\vtheta}(\vy_1) - \vy_2}^2} \\
    &= \argmin_{\vtheta} \EX{\norm[2]{f_{\vtheta}(\vy_1) - \vx - \ve_2 }^2} \\
    &= \argmin_{\vtheta}\EX{\norm[2]{f_{\vtheta}(\vy_1)}^2 - 2 \transp{\vx} f_{\vtheta}(\vy_1) - 2 \transp{\ve_2} f_{\vtheta}(\vy_1)} \\
    &= \argmin_{\vtheta}\EX{\norm[2]{f_{\vtheta}(\vy_1)}^2 - 2 \transp{\vx} f_{\vtheta}(\vy_1)}\\
    &= \vtheta_{\mathrm{N2C}},    
\end{align*}

 which concludes the proof. Here, the second to last equality follows from the noise being independent and having zero mean.

 \section{Sample Reconstructions}

\begin{figure}[h]
  \captionsetup[subfigure]{labelformat=empty,justification=centering}
  \centering
  \subfloat[Ground Truth]{\includegraphics[width=0.14\textwidth]{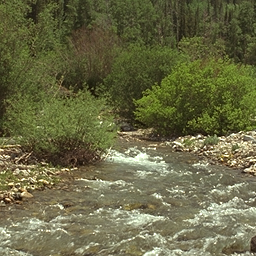}}
    \hfill
   \subfloat[ Noisy 28.1 dB]{\includegraphics[width=0.14\textwidth]{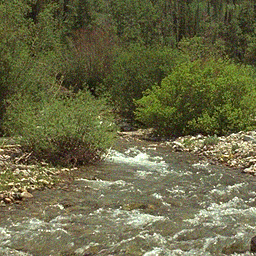}}
    \hfill
    \subfloat[ BM3D 29.9 dB]{\includegraphics[width=0.14\textwidth]{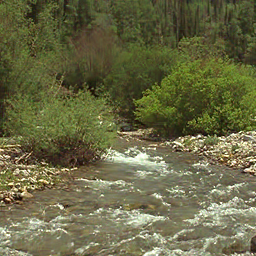}}
    \hfill  
    \subfloat[ S2S 25.2 dB]{\includegraphics[width=0.14\textwidth]{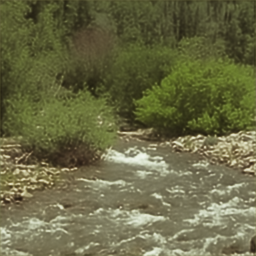}}
    \hfill 
    \subfloat[ S2S* 23.2 dB]{\includegraphics[width=0.14\textwidth]{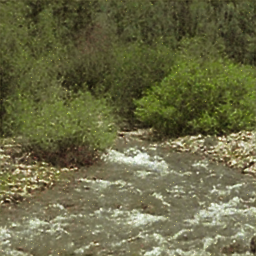}}
    \hfill 
     \subfloat[ DIP 30.2 dB]{\includegraphics[width=0.14\textwidth]{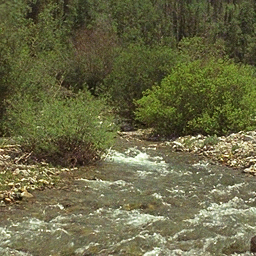}}
    \hfill    
  \subfloat[ Ours 30.2 dB]{\includegraphics[width=0.14\textwidth]{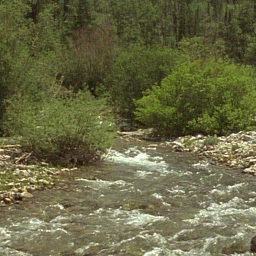}}
  \vfill
  
  \subfloat[Ground Truth]{\includegraphics[width=0.14\textwidth]{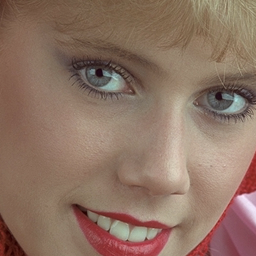}}
    \hfill
   \subfloat[ Noisy 20.2 dB]{\includegraphics[width=0.14\textwidth]{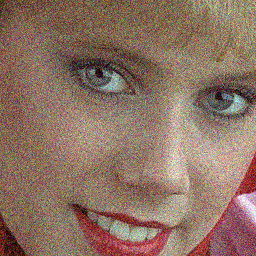}}
    \hfill
    \subfloat[ BM3D 32.6 dB]{\includegraphics[width=0.14\textwidth]{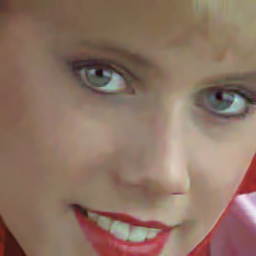}}
    \hfill  
    \subfloat[ S2S 33.0 dB]{\includegraphics[width=0.14\textwidth]{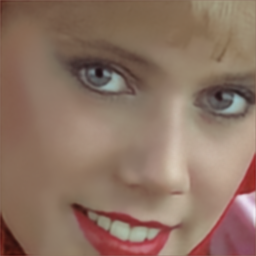}}
    \hfill 
    \subfloat[ S2S* 30.9 dB]{\includegraphics[width=0.14\textwidth]{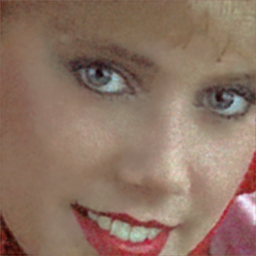}}
    \hfill 
     \subfloat[ DIP 30.8 dB]{\includegraphics[width=0.14\textwidth]{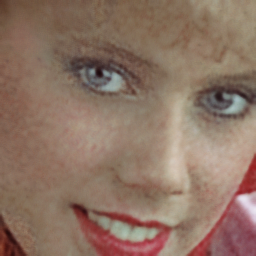}}
    \hfill    
  \subfloat[ Ours 31.6 dB]{\includegraphics[width=0.14\textwidth]{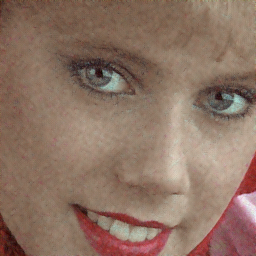}}
  \vfill
  
    \subfloat[Ground Truth]{\includegraphics[width=0.14\textwidth]{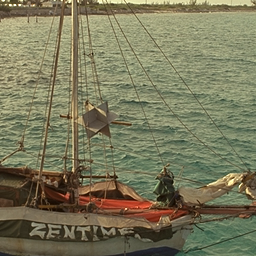}}
    \hfill
   \subfloat[ Noisy 14.5 dB]{\includegraphics[width=0.14\textwidth]{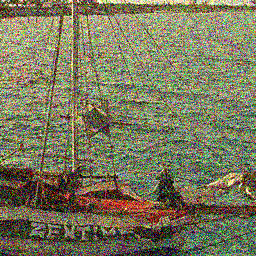}}
    \hfill
    \subfloat[ BM3D 22.8 dB]{\includegraphics[width=0.14\textwidth]{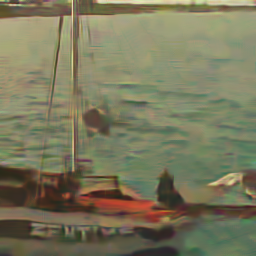}}
    \hfill  
    \subfloat[ S2S 23.8 dB]{\includegraphics[width=0.14\textwidth]{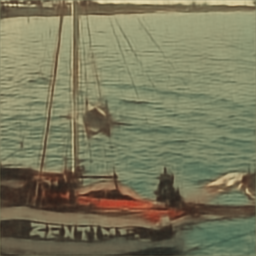}}
    \hfill 
    \subfloat[ S2S* 22.9 dB]{\includegraphics[width=0.14\textwidth]{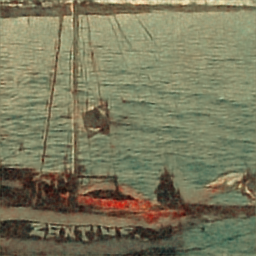}}
    \hfill 
     \subfloat[ DIP 21.9 dB]{\includegraphics[width=0.14\textwidth]{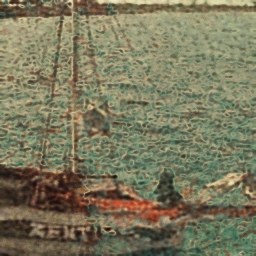}}
    \hfill    
  \subfloat[ Ours 23.4 dB]{\includegraphics[width=0.14\textwidth]{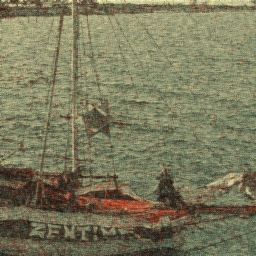}}
  \caption{Gaussian denoising on Kodak24 images. Upper row: $\sigma=10$, middle row: $\sigma=25$, lower row; $\sigma=50$. Note how Self2Self fails on the low noise level (top row, $\sigma=10$), and produces an image noisier than the input noisy image.}
\end{figure}

\begin{figure}[h]
  \captionsetup[subfigure]{labelformat=empty,justification=centering}
  \centering
  \subfloat[Ground Truth]{\includegraphics[width=0.14\textwidth]{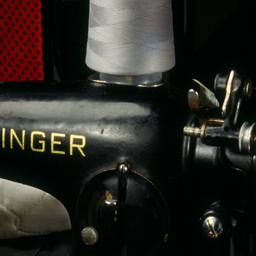}}
    \hfill
   \subfloat[ Noisy 24.9 dB]{\includegraphics[width=0.14\textwidth]{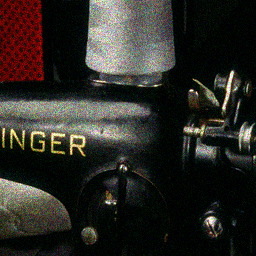}}
    \hfill
    \subfloat[ BM3D 28.4 dB]{\includegraphics[width=0.14\textwidth]{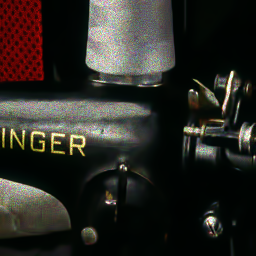}}
    \hfill  
    \subfloat[ S2S 33.0 dB]{\includegraphics[width=0.14\textwidth]{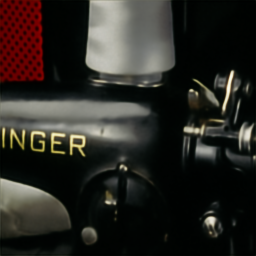}}
    \hfill 
    \subfloat[ S2S* 29.9 dB]{\includegraphics[width=0.14\textwidth]{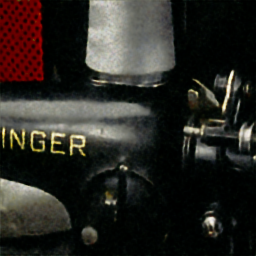}}
    \hfill 
     \subfloat[ DIP 28.0 dB]{\includegraphics[width=0.14\textwidth]{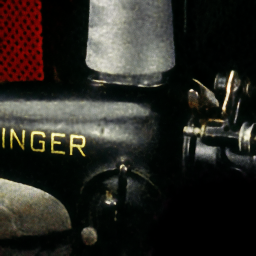}}
    \hfill    
  \subfloat[ Ours 33.1 dB]{\includegraphics[width=0.14\textwidth]{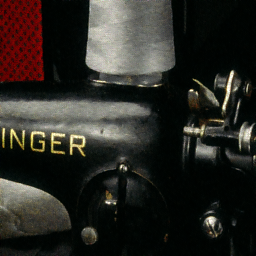}}
  \vfill
  
    \subfloat[Ground Truth]{\includegraphics[width=0.14\textwidth]{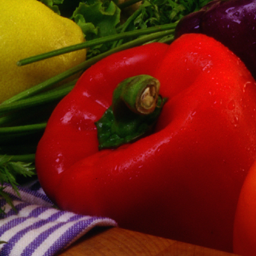}}
    \hfill
   \subfloat[ Noisy 21.0 dB]{\includegraphics[width=0.14\textwidth]{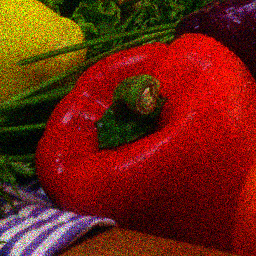}}
    \hfill
    \subfloat[ BM3D 23.0 dB]{\includegraphics[width=0.14\textwidth]{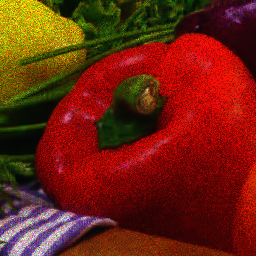}}
    \hfill  
    \subfloat[ S2S 32.7 dB]{\includegraphics[width=0.14\textwidth]{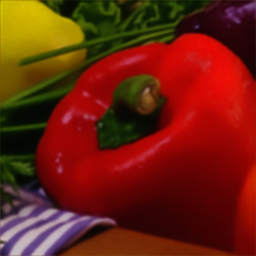}}
    \hfill 
    \subfloat[ S2S* 29.6  dB]{\includegraphics[width=0.14\textwidth]{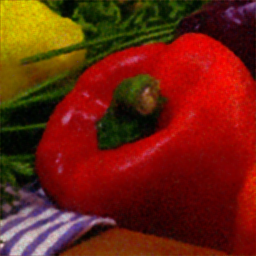}}
    \hfill 
     \subfloat[ DIP 29.7 dB]{\includegraphics[width=0.14\textwidth]{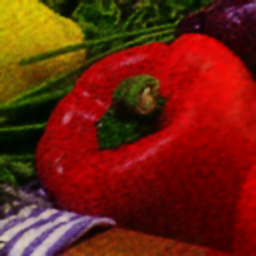}}
    \hfill    
  \subfloat[ Ours 30.8 dB]{\includegraphics[width=0.14\textwidth]{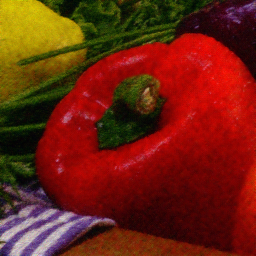}}
  \vfill
  
    \subfloat[Ground Truth]{\includegraphics[width=0.14\textwidth]{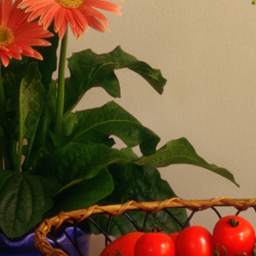}}
    \hfill
   \subfloat[ Noisy 15.4 dB]{\includegraphics[width=0.14\textwidth]{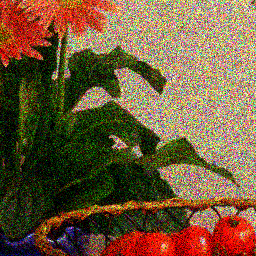}}
    \hfill
    \subfloat[ BM3D 25.0 dB]{\includegraphics[width=0.14\textwidth]{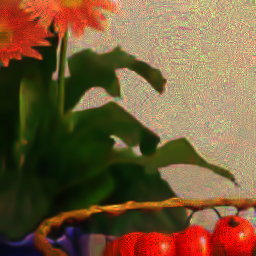}}
    \hfill  
    \subfloat[ S2S 30.0  dB]{\includegraphics[width=0.14\textwidth]{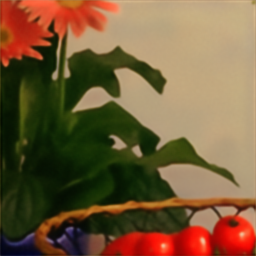}}
    \hfill 
    \subfloat[ S2S* 27.7 dB]{\includegraphics[width=0.14\textwidth]{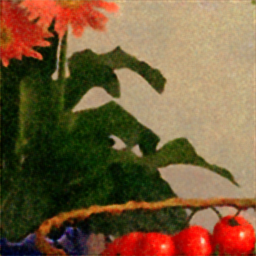}}
    \hfill 
     \subfloat[ DIP 24.9 dB]{\includegraphics[width=0.14\textwidth]{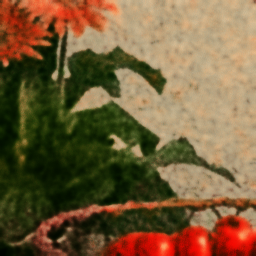}}
    \hfill    
  \subfloat[ Ours 27.3 dB]{\includegraphics[width=0.14\textwidth]{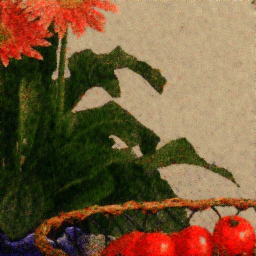}}
  \vfill
  \caption{Poisson denoising on McMaster18 images. Upper row: $\lambda=50$, middle row: $\lambda=25$, lower row; $\lambda=10$. Note the inconsistency of BM3D's performance, as it relies on noise level estimation, which varies from image to image.}

\end{figure}

\end{document}